\pdfoutput=1
\documentclass[11pt]{article}

\usepackage{acl}

\usepackage{times}
\usepackage{latexsym}
\usepackage{todonotes}
\usepackage{tabu}
\usepackage{booktabs}
\usepackage{amsfonts, bm, amsmath}
\usepackage{multirow, tabularx}
\usepackage{times}
\usepackage{latexsym}
\usepackage[T1]{fontenc}
\usepackage[utf8]{inputenc}
\usepackage{microtype}
\usepackage{arydshln}
\usepackage[capitalise]{cleveref}

\usepackage{microtype}
\newcommand\blfootnote[1]{%
  \begingroup
  \renewcommand\thefootnote{}\footnote{#1}%
  \addtocounter{footnote}{-1}%
  \endgroup
}

\title{Attention-guided Generative Models for Extractive Question Answering}

\author{Peng Xu$^*$\quad Davis Liang$^*$\quad Zhiheng Huang\quad Bing Xiang \\ \texttt{\{pengx, liadavis, zhiheng, bxiang\}@amazon.com} \\ \\ AWS AI Labs}

\begin{document}
\maketitle
\begin{abstract}
We propose a novel method for applying Transformer models to extractive question answering (QA) tasks. Recently, pretrained generative sequence-to-sequence (seq2seq) models have achieved great success in question answering. Contributing to the success of these models are internal attention mechanisms such as cross-attention. We propose a simple strategy to obtain an extractive answer span from the generative model by leveraging the decoder cross-attention patterns. Viewing cross-attention as an architectural prior, we apply joint training to further improve QA performance. Empirical results show that on open-domain question answering datasets like NaturalQuestions and TriviaQA, our method approaches state-of-the-art performance on both generative and extractive inference, all while using much fewer parameters. Furthermore, this strategy allows us to perform hallucination-free inference while conferring significant improvements to the model's ability to rerank relevant passages.
\end{abstract}
\blfootnote{$^*$ Corresponding Authors}
\section{Introduction}

Recently, it has been shown that pretrained generative seq2seq models can achieve great performance on question answering (QA) tasks \cite{raffel2020t5,roberts2020much,izacard2020leveraging}. 
In the Transformer \cite{vaswani2017attention} architecture, cross-attention plays a crucial role in extracting information from the encoder for use during generation. For the QA task, cross-attention serves as a natural architectural prior, aligning the generated answer tokens to evidence from within the context passages.
We posit that this alignment can be exploited as a means for answer span extraction.

Initially, we found that training a transformer model on the task of generative question answering and then leveraging the cross-attention patterns to perform zero-shot answer span extraction works remarkably well, with results approaching the generative baseline. Consequently, we propose several additional joint training strategies that further improve both the generative and extractive QA performance. 

\begin{figure*}[t!]
  \centering\includegraphics[width=1.0\linewidth]{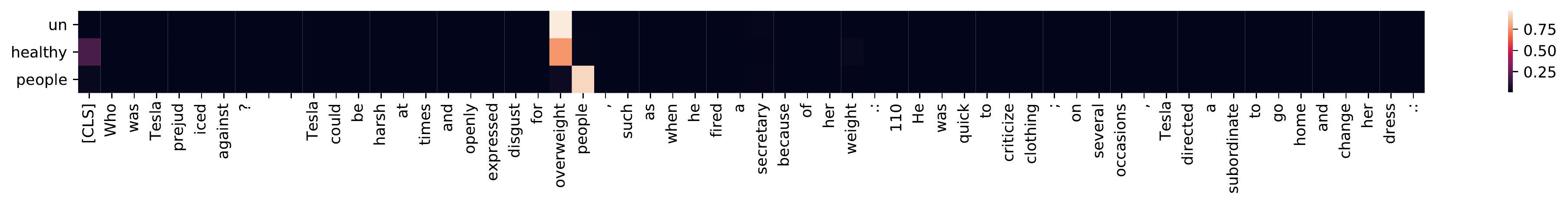}
  \centering\includegraphics[width=1.0\linewidth]{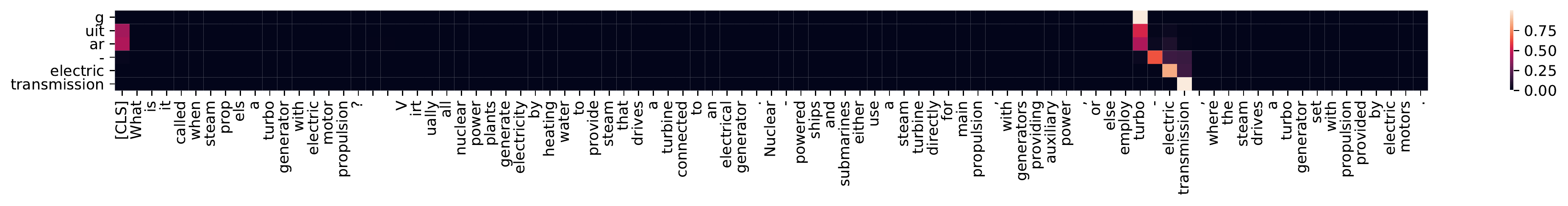}
  \centering\includegraphics[width=1.0\linewidth]{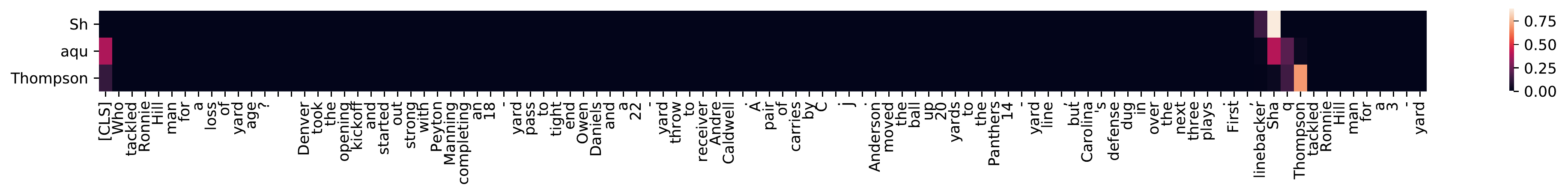}
  \centering\includegraphics[width=1.0\linewidth]{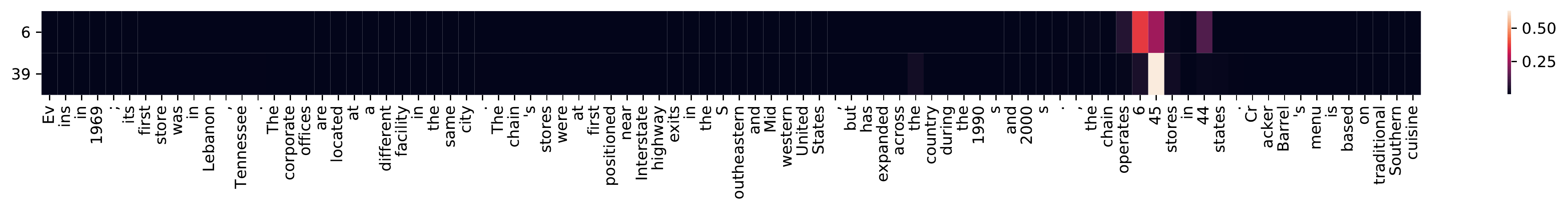}
  \caption{All examples shown are taken from the NaturalQuestions dataset. (Row 1): the decoder predicts `unhealthy people' when the actual answer, `overweight people', is correctly extracted from the cross-attention alignments. (Row 2): the decoder generates `guitar-electric transmission' when the cross-attention was aligned to the correct answer, `turbo-electric transmission'. (Row 3): the decoder misspells `Shaq Thompson', possibly in an attempt to generate `Shaquille'; nonetheless, the cross-attention alignments correctly answer this question. (Row 4): This example, which corresponds to the question \emph{``how many cracker barrels in the united states''}, presents an example of hallucinations that can arise through memorization of outdated information. Natural Questions is known to have test-set leakage \cite{lewis2020question} and this example occurs in the training set with an answer of `639' (extracted from the 2017 dump of Wikipedia) but the cross-attention is aligned to `645' in the new context, which was retrieved from the 2018 dump. In this example, the first 80 tokens of context are dropped for readability.}  \label{fig:hallucination}
\end{figure*}

Additionally, we point out that generative models are notoriously brittle and prone to hallucination. In particular, out-of-context predictions can cause significant degradations in QA performance. We show a few examples in \cref{fig:hallucination}. Generative models incur an additional risk factor in industry question answering systems, where hallucinations are especially hazardous. To address this issue, we show that our cross-attention extracted spans can also be leveraged as a fallback option to prevent hallucinations as they occur.

Finally, we demonstrate that our models, trained with the joint generative-extractive objective, offers significant improvements over previous baselines in passage ranking.

\begin{figure*}[t!]
  \centering\includegraphics[width=1.0\linewidth]{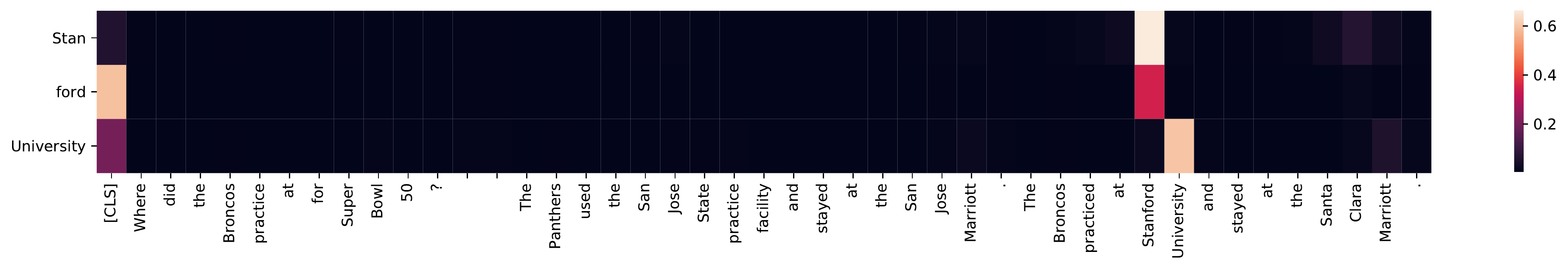}
  \caption{Average cross-attention between generated tokens and input context tokens on an example from SQuAD-v1.1 dataset. Each row and column corresponds to a decoder generated token and an input token, respectively. The generated token sequence [`Stan', `ford', `University'] attends to the input tokens [`Stanford', `University'].}
  \label{fig:stanford}
\end{figure*}

Our contributions can be summarized as follows:
\begin{itemize}
    \item We propose a simple attention-based inference strategy to extract answer spans from a seq2seq Transformer model without introducing any additional parameters. 
    \item We propose a joint training strategy by combining the normal generative loss and a span extractive loss via enforcing the cross-attention to align with answer span positions within the context passages. 
    \item We conduct an empirical study on both closed-domain and open-domain question answering tasks. We show that our methods achieve performance similar to the state-of-the-art using much fewer parameters.
    \item We propose additional ways to use the cross-attention probabilities (1) to leverage the extractive predictions as a fallback to help prevent hallucination and (2) as a proxy for passage scores to help provide additional context for the generated answer, improving the interpretability of an otherwise black-box model. 
\end{itemize}
\section{Method}
In this section, we first give a brief overview of the cross-attention mechanism in the Transformer models \cite{vaswani2017attention} in \cref{sec:cross_attention}. Then we describe how to use cross attention for span extractions in \cref{sec:inference}. Finally, we introduce the joint training strategy for combining the generative and extractive losses in \cref{sec:jointraining}.

\subsection{Cross-attention in transformer models}\label{sec:cross_attention}
We briefly describe the cross-attention mechanism used in the seq2seq Transformer model \cite{vaswani2017attention}. 
In the Transformer architecture, the encoder will produce a global contextual representation, $\mathbf X\in \mathbb R^{n\times d}$, for a sequence of $n$ input tokens, and the decoder will attend on the encoder representations to generate each token of the final predicted sequence. Specifically, let $\mathbf h \in \mathbb R^{d}$ be the output hidden representation from the self-attention layer in the decoder at a certain position. The cross-attention mechanism will perform the following operations. First, it computes the query, key and value vectors through a linear layer with weights $W_q, W_k, W_v$,
\begin{align*}
    \mathbf q = \mathbf h W_q, ~~ \mathbf K = \mathbf X W_k, ~~ \mathbf V = \mathbf X W_v.
\end{align*}
Then, it computes the similarity scores between the query and key vectors at different positions. The scores are normalized via softmax across all the input token positions, also known as the attention weights.
\begin{align*}
    \alpha_i = \mathbf q^T \mathbf K_i, ~~ \tilde\alpha_i = \dfrac{\exp(\alpha_i)}{\sum_{j=1}^n \exp(\alpha_j)}.
\end{align*}
Finally it computes the updated representation by aggregating the value vectors using the attention weights, followed by a linear layer with weights $W_o$,
\begin{align}
\tilde{\mathbf h} = \left(\sum_{i=1}^n \tilde\alpha_i \mathbf V_i\right) W_o.
\label{eq:joint}
\end{align}
A typical Transformer model usually consists of multiple cross-attention modules in parallel (multi-head attention) with normalization layers and skip connections. Please refer to \citet{vaswani2017attention} for additional details.

\subsection{Span extraction via cross-attention pattern}\label{sec:inference}
In this subsection, we describe how to use cross-attention patterns to perform answer span extraction. Specifically, given a question $q$ and a context passage $C=[x_1, x_2, \ldots, x_n]$, the model will take $[q \texttt{[SEP]} C]$ as input and autoregressively generate an answer $\hat a = [\hat y_1, \ldots, \hat y_t]$. 
Intuitively, the cross-attention weights illustrate the importance of each token in the source input in each decoding step. In the case of reading comprehension, the more attended on a context token is, the more relevant that token is to the corresponding generated answer token. Under this assumption, we propose a simple strategy to extract answer spans while decoding.

More concretely, we examine the cross-attention weights of the last decoder layer when the model generates the \textit{first} token $\hat y_1$ and the \textit{last} answer token $\hat y_t$. We hypothesize that when the model decodes the first answer token, the most attended position is likely the start position of the extractive answer span. Similarly, when model decodes the last token, the most attended position is likely the end position of the answer span. We consider the cross-attention weights when decoding the first and last answer tokens as a reasonable proxy for the probabilities of the start and end positions of the answer span, i.e.
\begin{equation}
\begin{aligned}
    \Pr[start=i] &= CrossAttn(\hat y_1, x_i), \\
    \Pr[end=i] &= CrossAttn(\hat y_t, x_i).
\end{aligned}
\label{eq:span_inf}
\end{equation}
where $CrossAttn()$ is a function that returns the cross-attention probability between two tokens.

\begin{figure*}[t!]
  \centering\includegraphics[width=1.0\linewidth]{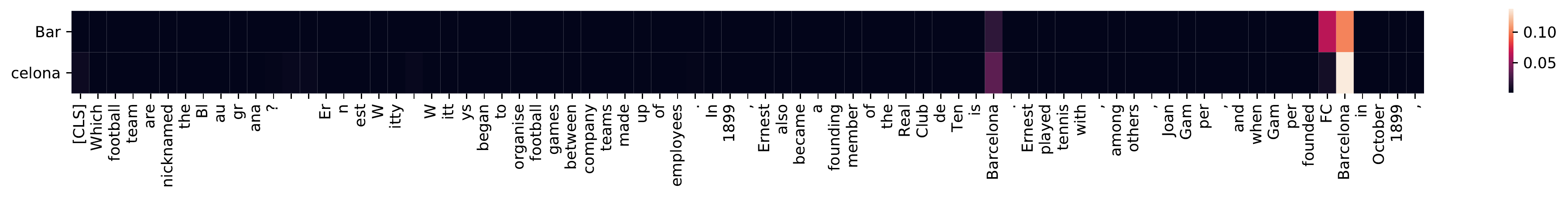}
  \centering\includegraphics[width=1.0\linewidth]{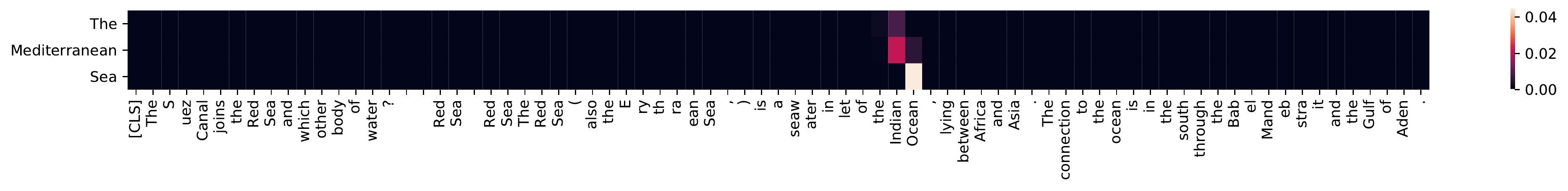}
  \centering\includegraphics[width=1.0\linewidth]{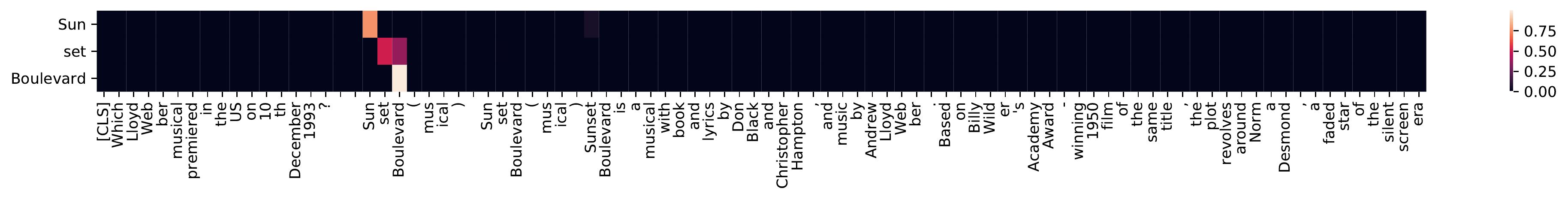}
  \caption{All examples from TriviaQA. (Top): The highly attended span `FC Barcelona', a football club, is the correct answer. The other Barcelona refers to the `Real Club Tennis Barcelona' which is a Tennis club. (Middle): if no possible answers are found, the model may attempt to highlight the most plausible one, if one exists. (Bottom): If there are multiple correct occurrences of the answer, the model will generally highlight the first one.}  \label{fig:misc_cases}
\end{figure*}

\begin{figure*}[t!]
  \centering\includegraphics[width=1.0\linewidth]{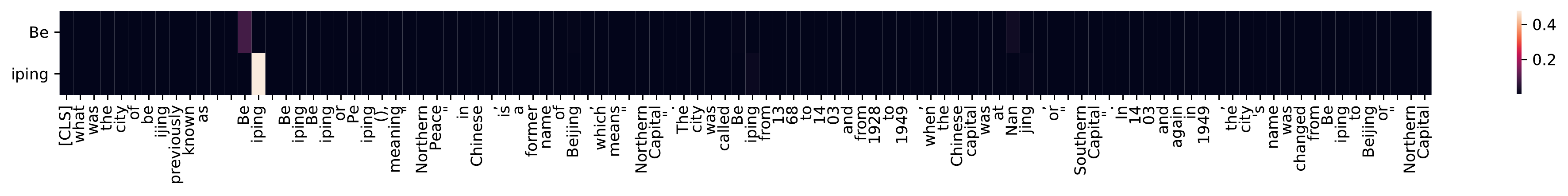}
  \centering\includegraphics[width=1.0\linewidth]{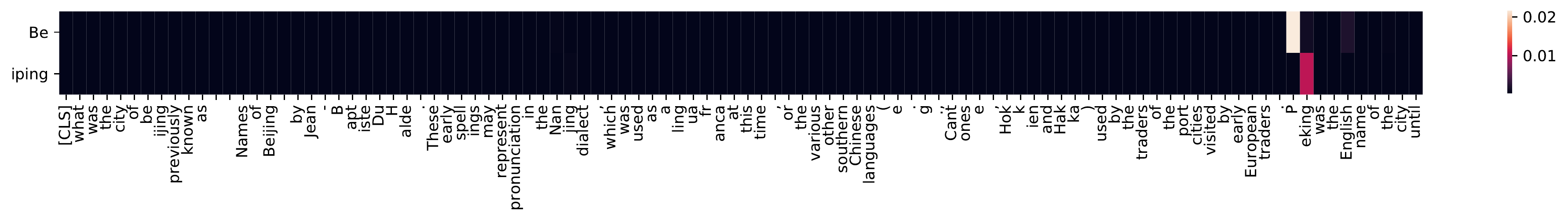}
  \centering\includegraphics[width=1.0\linewidth]{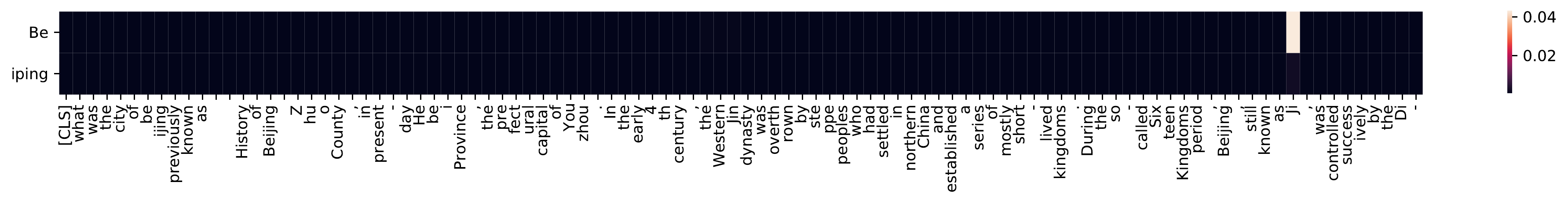}
  \centering\includegraphics[width=1.0\linewidth]{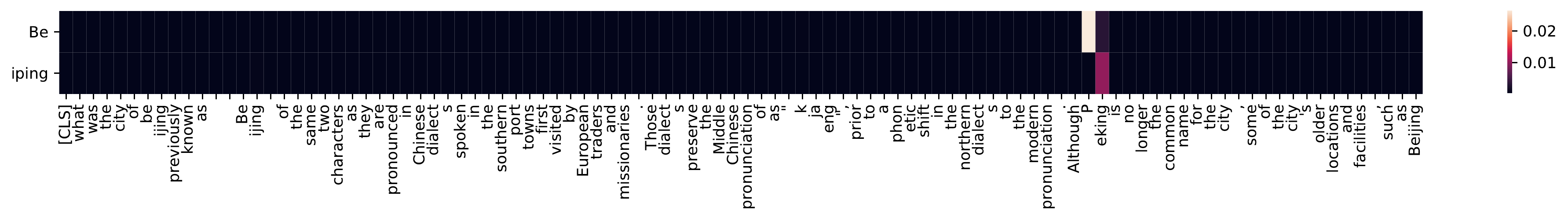}
  \caption{This example, which corresponds to the question \textit{``what was the city of beijing previously known as''}, presents the interesting situation where different correct answers exist across multiple passages. The model distributes the probability across multiple answers (`Beiping', `Peking', and `Ji') from different passages, putting a significant portion of of the probability mass on the most likely prediction.}  \label{fig:beijing}
\end{figure*}

To account for the multi-head attention in Transformers, we simply take the average cross-attention probabilities over multiple heads. In addition, Transformer models can be easily extended to the class of Fusion-in-Decoder(FiD) models to deal with multiple source inputs \cite{izacard2020leveraging}. Since the cross-attention mechanism remains the same in FiD models, the methodology described above still applies. Visualization of the cross-attention alignments and examples of how they behave in various situations are shown in \cref{fig:stanford},\ref{fig:hallucination}, \ref{fig:misc_cases}, and \ref{fig:beijing}.

\subsection{Joint generative and extractive training}\label{sec:jointraining}
To further improve the cross-attention patterns for span extraction, we incorporate a span extraction loss into the training process.  In the normal seq2seq model training, the generative loss is minimized through teacher-forcing, i.e.
\begin{align*}
    \ell_{gen}(q, C, a) = - \sum_{i=1}^t \log \Pr\left(y_i |y_{1\ldots, i-1}; q,C\right)
\end{align*}
where $a=[y_1,\ldots, y_t]$ is the ground truth answer. Based on \cref{eq:span_inf}, we can obtain the typical span extraction loss via the cross-entropy loss of the start and end position predictions, i.e.
\begin{align*}
    \ell_{span}(q, C, start, end) = \quad\quad\quad\quad\quad\quad\quad\quad\quad\quad\\
    CE(start, CrossAttn(y_1, C)) \\
    + ~CE(end, CrossAttn(y_t, C)),
\end{align*}
where $start, end$ are the start and end positions of the ground truth answer span, respectively, and $CE()$ computes the cross-entropy loss. Then, the final joint training loss function is as follows:
\begin{align*}
    \ell_{joint}(q, C, a) = (1 - \lambda) \ell_{gen} + \lambda \ell_{span},
\end{align*}
where $\lambda \in [0, 1]$ is a hyperparameter.


\section{Experiments and results}
In this section, we apply our proposed methods on the reading comprehension and open-domain question answering tasks. Specifically, we initialize our FID models with the pretrained BART-base and BART-large parameters \cite{lewis2019bart} in all of our experiments. 

\subsection{Reading comprehension}
We conduct an empirical evaluation of our models on the reading comprehension dataset SQuAD-v1.1. At inference time, we use a beam size of 5. While autoregressively decoding, we also extract answer spans based on the cross-attention weights, as previously described. The results are shown in \cref{table:closed_domain}.

\begin{table}[h!]
\begin{minipage}{1.0\linewidth}
	\centering
	\footnotesize

\begin{tabu}{@{}llccccc@{}}
\toprule
 &  \multicolumn{4}{c}{\textsc{SQuAD-v1.1}} \\
  &   \multicolumn{2}{c}{\textsc{Generative}} & \multicolumn{2}{c}{\textsc{Extractive}} \\
\cmidrule(r){2-3} \cmidrule(r){4-5} 
\textbf{Model} & EM & F1 & EM & F1 \\
\midrule
T5-base (\textsc{220M}) & 85.44 & 92.08 & - & - \\
T5-large (\textsc{770M}) & 86.66 & 93.79 & - & - \\
\midrule
BART-base (\textsc{139M}) & 79.96 & 88.14 & 79.99 & 88.07\\
BART-base joint   & 79.82 & 88.12& \textbf{81.13} & \textbf{88.85} \\
\hdashline
BART-large (\textsc{406M})  & 84.23 & 91.56 & 62.94 & 73.82 \\
BART-large joint  & 84.17 & 91.69& \textbf{85.53} & \textbf{92.41} \\
\bottomrule
\end{tabu}
\end{minipage}
\caption{Results on SQuAD-v1.1 show that our method improves over the BART baselines.The numbers in () are the number of model parameters.}
\label{table:closed_domain}
\end{table}

In \cref{table:closed_domain}, we find that BART-base trained only on the generative task has reasonable performance on extractive span prediction. This indicates that the cross-attention patterns have implicitly learned to extract answer spans through vanilla generative training with teacher-forcing. Explicitly enforcing this signal via the joint training objective \cref{eq:joint} obtains addition improvements to the extractive answer predictions, outperforming even the generative predictions. Note that for BART-large, training only on the generative loss produces lower quality extractive predictions. We hypothesize that compared to BART-base, the increased depth (and therefore increased capacity) of BART-large makes the model less reliant on the cross-attention alignments for answer generation.

\begin{table*}[t!]
\begin{minipage}{1.0\linewidth}
	\centering
	\footnotesize
\begin{tabu}{@{}ll|cccc|cccc@{}}
\toprule
& & \multicolumn{4}{c}{\textsc{NaturalQuestions}} & 
\multicolumn{4}{|c}{\textsc{TriviaQA}} \\
& & \multicolumn{2}{c}{\textsc{Generative}} & 
\multicolumn{2}{c}{\textsc{Extractive}} & \multicolumn{2}{|c}{\textsc{Generative}} & 
\multicolumn{2}{c}{\textsc{Extractive}} \\
\cmidrule(r){3-4} \cmidrule(r){5-6} \cmidrule(r){7-8} \cmidrule(r){9-10}
\textbf{Model} & \textbf{Params.} & \textbf{dev}  & \textbf{test}  & \textbf{dev}  & \textbf{test}  & \textbf{dev}  & \textbf{test}  & \textbf{dev}  & \textbf{test}  \\
\midrule

DPR\textit{(BERT-base)}\cite{karpukhin-etal-2020-dense} &  \textsc{110M} & - & - & - & 41.5 & - & - & - & 57.9 \\
ColBERT\textit{(BERT-base)}\cite{khattab2020relevance} &  \textsc{110M} & - & - & - & 42.5 & - & - & - & 63.2  \\
ColBERT\textit{(BERT-large)}\cite{khattab2020relevance} &  \textsc{330M} & - & - & - & 48.2 & - & - & - & - \\
\hdashline
RAG \textit{(BART-large)}\cite{lewis2020retrieval} &  \textsc{406M} & - & 44.5 & - & - & - & 56.8 & - & - \\
FiD\textit{(T5-base)}\cite{izacard2020leveraging}  &  \textsc{220M} & - & 48.2 & - & - & - & 65 & - & - \\
FiD\textit{(T5-large)}\cite{izacard2020leveraging}  &  \textsc{770M} & - & 51.4 & - & - & - & 67.6 & - & - \\
FiD-KD\textit{(T5-base)}\cite{izacard2020distilling} &  \textsc{220M} & 48.0 & 49.6 & - & - & 68.6 & 68.8 & - & - \\
FiD-KD\textit{(T5-large)}\cite{izacard2020distilling}  &  \textsc{770M} & 51.9 & \textbf{53.7} & - & - & 71.9 & \textbf{72.1} & - & -\\
\midrule
Our FiD \textit{(BART-base)} &  \textsc{139M} & 47.33 & 48.03 & 41.05      & 41.63    & 67.46  & 67.79  & 58.16  & 58.31                \\
Our FiD \textit{(BART-base)} + joint &  \textsc{139M} & \textbf{48.59} & \textbf{49.09} & \textbf{46.15}  & \textbf{46.18}  & \textbf{67.70} & \textbf{68.16} & \textbf{62.99} & \textbf{63.72}       \\
Our FiD \textit{(BART-large)} &  \textsc{406M} & 51.24 & 52.58 & 32.74      & 35.15    & 69.64  & 70.64  & 41.69  & 40.46                \\
Our FiD \textit{(BART-large)} + joint &  \textsc{406M} & \textbf{52.24} & \textbf{53.43} & \textbf{49.14}  & \textbf{50.03}  & \textbf{70.61} & \textbf{71.15} & \textbf{66.23} & \textbf{67.21}       \\
\bottomrule
\end{tabu}
\end{minipage}
\caption{Our model is able to generate and extract span-based answers without any additional parameters, achieving state-of-the-art results in many settings. Note that our largest model, BART-large, is approximately half the size of the SOTA T5-large model.}
\label{table:open_domain}
\end{table*}

\subsection{Open-domain question answering}
In this section, we conduct an empirical evaluation on open-domain question answering (ODQA) tasks. A typical ODQA system consists of two steps. First, for each question, we retrieve supporting passages from a large external corpus. Then, a reader model takes the retrieved passages to predict the answer. In this work, we focus on the reader model only. 
Specifically, we consider the Fusion-in-Decoder (FiD) architecture initialized from the BART models. 

We perform experiments on the NaturalQuestions \cite{kwiatkowski2019natural} and TriviaQA \cite{JoshiTriviaQA2017} datasets, using same settings as the original FiD paper \cite{izacard2020distilling,izacard2020leveraging}. We also use the same dataset partitions; NaturalQuestions has a train/dev/test split of 79,168 / 8,757 / 3,610 queries and TriviaQA has a train/dev/test split of 78,785 / 8,837 / 11,313 queries. For fair comparison, we directly borrow the retrieved passages from \cite{izacard2020distilling} instead of performing retrieval ourselves. We report the standard Exact Match (EM) metric on the final answer predictions. 

\paragraph{Training} We apply distant supervision to train our reader models. Since there are no human annotated \textit{start} and \textit{end} positions in the training set, the ground truth answer can potentially be found in multiple retrieved passages. 
To account for this, we consider three strategies. (1) \textit{Multi-label spans}: Consider all the occurrences and use multi-label cross-entropy loss. (2) \textit{First span}: Take the first occurrence as the the start and end positions, assuming the top passages are more relevant. (3) \textit{Most-likely span}: Consider the span occurrence with the \textit{highest likelihood} based on the attention scores. We find that on the datasets provided by \cite{izacard2020distilling}, the \textit{First span} strategy works the best. We make detailed comparisons in \cref{sec:training_exp}.

\paragraph{Inference} For generative answer prediction, we apply beam search decoding and choose the best beam. For extractive span prediction, we consider a simple local greedy approach. Given the start and end position probabilities $P_{start}, P_{end}$ (based on cross-attention weights of the first and last decoded tokens), we consider two span prediction candidates,
\begin{align*}
\centering
&
\begin{cases}
    \hat{start}_1 &= \arg\max P_{start}, \\
    \hat{end}_1 &= \arg\max_{\hat{start}_1<i < \hat{start}_1+ l_{max}} P_{end}
\end{cases}
\\
&
\begin{cases}
    \hat{end}_2 &= \arg\max P_{end}, \\
    \hat{start}_2 &= \arg\max_{\hat{end}_2 - l_{max} <i < \hat{end}_2} P_{start}
\end{cases}
\end{align*}
where $l_{max}$ is the maximum answer span length. 
The final span prediction is the span with the larger probability among these two candidates.

\paragraph{Technical details}
We utilize the retrieved results\footnote{https://github.com/facebookresearch/FiD/blob/master/get-data.sh} from \citet{izacard2020distilling}, which contain 100 retrieved passages for each question. We finetune the FiD models initialized with the pretrained BART models on each dataset independently, using Adam \cite{kingma2014adam} with a learning rate of 5e-5 and warmup rate of 0.1. We train the models for 10 epochs with batch size 64 and validate at each epoch. The hyperparameter $\lambda$ is selected through cross validation.

\subsubsection{Main results}
We compare our models with state-of-the-art ODQA systems. The main results are shown in \cref{table:open_domain}. First, we notice that similar to the RC task, the cross-attention pattern produces impressive zero-shot results for extractive span prediction in the open-domain and that joint training can further improve the performance. On the NQ test set, our best model (BART-large with joint training) achieves 53.43 EM using the generative predictions and 50.03 EM using the extractive predictions; the generative prediction performance approaches the SOTA performance from FiD-KD, which uses a much larger model (T5-large), and the extractive prediction performance significantly outperforms the previous state-of-the-art extractive model (ColBERT). On TriviaQA, our best model underperforms FiD-KD (T5-large) in terms of generative prediction but achieves the state-of-the-art extractive performance.

\subsubsection{Ablation on training strategies}
\label{sec:training_exp}

\begin{table*}[ht]
\centering
\footnotesize
\begin{tabular}{lcccc}
\toprule
\multirow{2}{*}{\textbf{Model}} & \multirow{2}{*}{\textsc{Generative}} & \multicolumn{3}{c}{\textsc{Extractive}}               \\
                       &                                      & \textsc{Attention} & \textsc{Drop} & \textsc{Backoff} \\
\midrule
& \multicolumn{4}{c}{\textsc{NaturalQuestions}} \\
& \multicolumn{4}{c}{\textsc{(3,236 effective queries)}} \\
\cmidrule{2-5}
FiD(BART-base)              & 53.34                                & 46.38         & 53.12              & 53.65            \\
FiD(BART-base) joint        & 54.51                                & 51.42         & 54.39              & \textbf{54.82}   \\
FiD(BART-large)             & 58.25                                & 39.15         & 58.00              & 58.59            \\
FiD(BART-large) joint       & 59.18                                & 55.72         & 58.87              & \textbf{59.55}  \\
\midrule
& \multicolumn{4}{c}{\textsc{TriviaQA}} \\
& \multicolumn{4}{c}{\textsc{(9,966 effective queries)}} \\
\cmidrule{2-5}
FiD(BART-base) & 76.35 & 66.05 & 73.51 & 75.84 \\
FiD(BART-base) joint & 76.89 & 72.18 & 74.48 & \textbf{77.09} \\
FiD(BART-large) & 79.42 & 45.78 & 76.28 & 77.81 \\
FiD(BART-large) joint & 80.03 & 76.10 & 76.60 & \textbf{80.52} \\
\bottomrule
\end{tabular}
\caption{We benchmark various strategies to mitigate hallucinations. Results are reported on the test queries where the retrieved passages contain the ground truth answer. The numbers below the dataset names are the effective number of queries for each dataset. \textsc{Attention} uses the cross-attention to extract answer spans, \textsc{Drop} replaces all hallucinations with an empty string, and \textsc{Backoff} falls back to the extractive answer only when the generative prediction is a hallucination. The metric recorded is Exact Match (EM), in percent.}
\label{tab:hallucination}
\end{table*}

\begin{figure}[htp]
    \centering
    \includegraphics[width=0.49\textwidth]{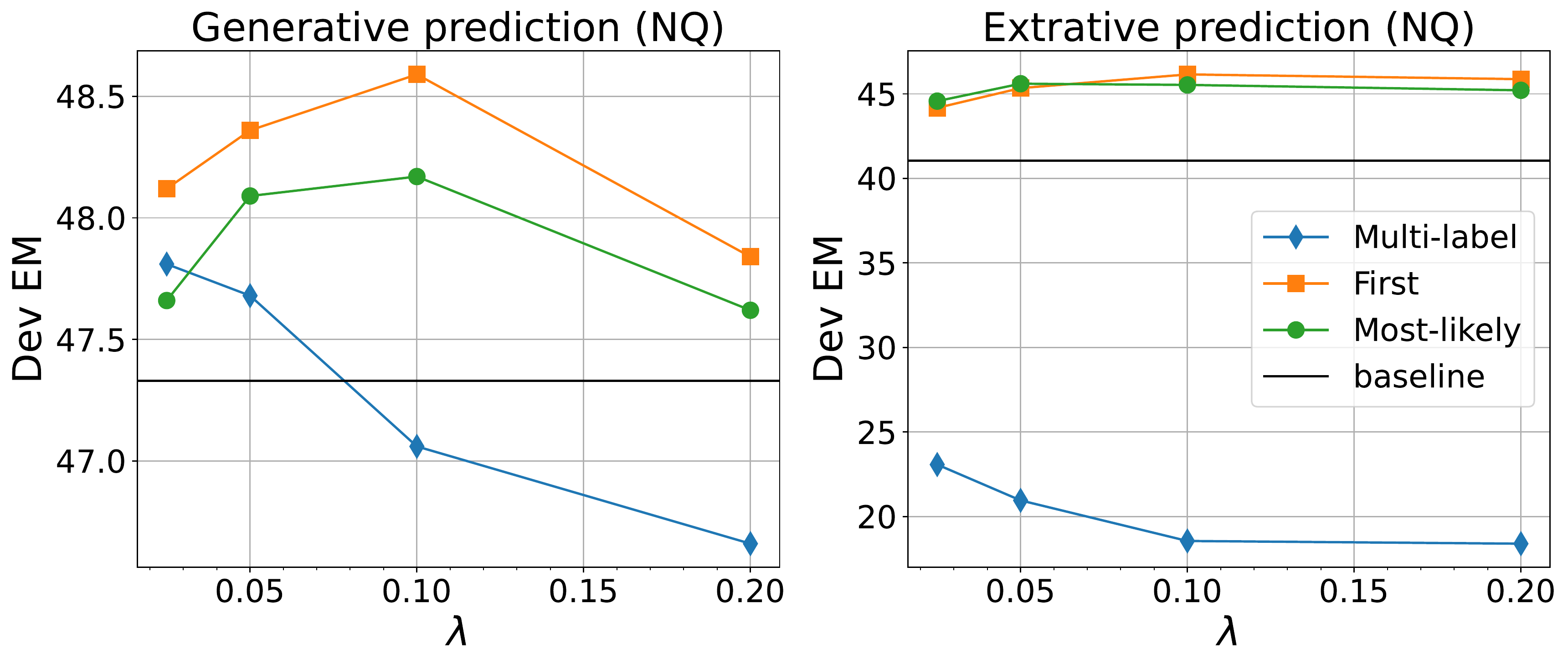}
    \caption{The dev set performance of various training strategies with the FiD (BART-base) model on the NaturalQuestions dataset. The baseline corresponds to the FiD model without joint training, i.e. $\lambda=0$, signified by the solid, horizontal reference line.}
    \label{fig:train}
\end{figure}

Here, we finetune the FiD (BART-base) model on NaturalQuestions with the three strategies described earlier. The results are shown in \cref{fig:train}. As we can see, the \textit{Multi-label Span} strategy clearly underperforms the two other approaches in both generative and extractive predictions. Increasing $\lambda$ degrades the performance further, below baseline. This result suggests that using every possible answer span as a training signal is extremely noisy (for example, some answer spans could be multi-sense words or be surrounded by the wrong context).

\textit{First Span} slightly outperforms the \textit{Most-likely Span} approach, suggesting that the ranking of the passages retrieved in \citet{izacard2020distilling} are relatively high quality. Nevertheless, we believe that the \textit{Most-likely Span} is a competitive and robust approach especially when well-ranked passages are unavailable.

\subsubsection{Reducing hallucination}
To further demonstrate the advantages of our approach, we consider the issue of hallucination in generative models. For example, on the NaturalQuestions test set, $6.2\%$ of generated answer predictions from a FiD (BART-large) model are not spans from within the 100 context passages. Similarly, on the TriviaQA test set, the rate of hallucination is even higher, at $11.2\%$. Please refer to \cref{fig:hallucination} for some real examples of the hallucinations the model generates.

To address this, we propose and compare various strategies for removing hallucinations. Our first method uses the attention-based extractive inference as a hallucination-free baseline (\textsc{Attention}). Secondly, we can simply drop out the out-of-context answers from the generative predictions (\textsc{Drop}). 

Finally, we can perform a simple \textit{backoff} which replaces hallucinations from the generative predictions with the extractive prediction (\textsc{Backoff}). To remove variations due to retrieval performance, we conduct the evaluation only on the queries where at least one of the ground truth answers exist in the retrieved passages. Under this constraint, we retain 3,236 test queries from NaturalQuestions and 9,966 test queries from TriviaQA. The results are shown in \cref{tab:hallucination}. We can see that (\textsc{Drop}) leads to a degradation of performance, especially for TriviaQA datasets (approx. 3 points EM). And using the simple \textsc{Backoff} strategy can effectively close this gap. For the models with joint training, \textsc{Backoff} achieves even better performance than the original generative predictions, indicating that \textsc{Backoff} is an effective \textit{hallucination-free} inference strategy for generative QA models. 

\subsubsection{Attention score as a reranker}
\label{sec:ranking}
In this section, we demonstrate that the cross-attention probabilities are a powerful proxy for the scores used in passage ranking. Particularly, in additional to answer prediction, the FiD reader model can also perform reranking over retrieved passages. Specifically, given a query $q$, one context passage $p_i$ among $C$, we define the passage score as follows:
\begin{equation}
\small
\begin{aligned}
    score(p_i| q) := \sum_{t \in [q;p_i]} P_{start}(t) \cdot  \sum_{t \in [q;p_i]} P_{end}(t)
\end{aligned}
\label{eq:pr}
\end{equation}
where $P_{start}$ and $P_{end}$ are the predicted start/end position probabilities of the context passages. 
\begin{table}[ht]
\begin{minipage}{1.0\linewidth}
	\centering
	\footnotesize
	\scriptsize
\begin{tabular}{lcccc}
\toprule
\textbf{Model} & \textbf{P@1} & \textbf{P@5} & \textbf{P@20} & \textbf{nDCG@20} \\
\midrule
& \multicolumn{4}{c}{\textsc{NaturalQuestions}} \\
\cmidrule{2-5}
FiD-KD (baseline) & 50.36 & 35.82 & 22.61 & 45.24 \\
\hdashline
FiD(BART-base) joint & 62.69 & 46.42 & 29.05 & 58.64 \\
FiD(BART-large) joint & \textbf{66.59} & \textbf{48.48} & \textbf{29.85} & \textbf{61.19} \\
\midrule
& \multicolumn{4}{c}{\textsc{TriviaQA}} \\
\cmidrule{2-5}
FiD-KD (baseline) & 55.43 & 46.84 & 38.57 & 45.94 \\
\hdashline
FiD(BART-base) joint & 65.33 & 57.97 & 49.97 & 60.29 \\
FiD(BART-large) joint & \textbf{67.44} & \textbf{59.62} & \textbf{51.05} & \textbf{62.17} \\
\bottomrule
\end{tabular}

\caption{Using cross attention scores as a proxy for passage scores. This allows us to effectively perform passage ranking using our FiD QA model. Results show that our joint generative-extractive training strategy can significantly improve our passage ranking results. All results are in percent.}
\label{tab:pr}
\end{minipage}
\end{table}

We conduct the passage ranking experiment based on the same passages provided by \citet{izacard2020distilling}. We assume that any passage that contains a ground truth answer as the relevant passages and the rest are considered irrelevant. We rerank the top 100 passages based on \cref{eq:pr} using our best FiD models. The results are shown in \cref{tab:pr}. We can see that the attention-based score defined by \cref{eq:pr} from our FiD models produce rankings that significantly improve upon the results from the neural retrieval model in \cite{izacard2020distilling}, which originally demonstrated the effectiveness of FiD as a passage reranker. Showing the reranked passages in additional to the answer prediction can provide additional context for the end-users or developers of ODQA systems to navigate. In practice, this can improve the interpretability and reliability of the typically black-box ODQA systems that we train and use today.



\section{Related work}

\paragraph{Extractive question answering} Extractive question answering problems have gained wide interest under both the closed-domain and open-domain settings. In the closed-domain setting, many methods have been proposed that continuously push the envelope on datasets such as SQuAD \cite{rajpurkar2016squad}, one of the most popular large-scale reading comprehension datasets. Many of the recent methods leverage pretrained language models with a simple token classification head to extract answer spans  \cite{devlin-etal-2019-bert,lan2019albert,liu2019roberta}. 

In the open-domain setting, \citet{chen-etal-2017-reading} introduced the retrieve-and-read framework to answer questions based on the unstructured Wikipedia corpus. Various works have been proposed to improve either or both retrieval and reader models \cite{wang-etal-2019-multi,karpukhin-etal-2020-dense,lee-etal-2019-latent,khattab2020relevance,lewis2020retrieval,min2019knowledge,izacard-grave-2021-leveraging, izacard2020distilling} under this framework.  

\paragraph{Generative models} Typically, generative models are used to tackle abstractive questions answering tasks, such as NarrativeQA \cite{kocisky-etal-2018-narrativeqa} and ELI5 \cite{fan-etal-2019-eli5}, where the answer usually does not appear as a span in the context. Recently, \cite{raffel2020t5} showed that large pretrained generative models can achieve competitive performance on SQuAD, an extractive question answering task. \citet{roberts2020much} proposed to use pretrained generative models to perform ODQA in a closed-book setting. 

More recently \citet{izacard2020leveraging} proposed the FiD models which significantly improved the end-to-end performance of ODQA. 
The issue of hallucination in generative models has recently gained attention in a variety of tasks including document summarization \cite{maynez-etal-2020-faithfulness}, machine translation \cite{zhou2020detecting}, news generation \cite{zeller2019defending}, and dialogue systems \cite{mielke2020linguistic,shuster2021retrieval}.

Our work is also inspired by the pointer network \cite{vinyals2015pointer, see-etal-2017-get}, where the cross-attention weights in a seq2seq model is used to replace or modify the distribution during decoding. In our case, we directly use the attention weights for extractive inference and provide additional useful context for generative predictions. Another closely related work is \citet{izacard2020distilling} which leverages the cross-attentions weights from FiD models to obtain weak supervision signals for training passage retrieval models, which aligns with our observation that attention scores can be used as a good proxy for passage scores (\cref{sec:ranking}).

\section{Conclusion}
Our work introduces a novel approach for using the cross-attention patterns of a generative QA model to obtain extractive answer spans. We propose methods to jointly train our model to perform generative and extractive inference, improving the performance of both. Furthermore, our method allows us to achieve hallucination-free inference while also improving the model's passage ranking capabilities. 

Our results demonstrate the effectiveness of leveraging cross-attention as a architectural prior for improving modern, state-of-the-art generative question answering systems. Our findings raise a number of questions to be explored next. In particular, while our empirical results show the effectiveness of strategies like \textsc{Backoff} for hallucination-free generation, we foresee the potential extension of using the cross-attention aligned tokens during the decoding process.  
\bibliography{anthology,custom}
\bibliographystyle{acl_natbib}

\end{document}